\PassOptionsToPackage{table}{xcolor}
\documentclass{bmvc2k}
\usepackage{bmvc2k_natbib} 

\usepackage{amsmath}
\usepackage{amssymb}
\usepackage{bm}
\usepackage{graphicx}

\usepackage{multirow}
\usepackage{booktabs}
\usepackage{siunitx}
\usepackage{wrapfig}
\usepackage{pgfplots}
\usepackage{xspace}
\usepackage{adjustbox}
\usepackage[normalem]{ulem}
\usepackage{caption}
\usepackage{algorithm}
\usepackage{algpseudocode}
\usepackage{placeins}
\usepackage{enumitem}

\definecolor{oursgreen}{RGB}{221,245,221}
\definecolor{rowgray}{RGB}{210,210,210}

\pgfplotsset{compat=1.16}
\usetikzlibrary{patterns}
\usepgfplotslibrary{colormaps}

\setlist{itemsep=0pt,parsep=0pt,topsep=0pt,partopsep=0pt}

\newcommand{\method}{\texttt{GATE}\xspace}

\title{GATE: Reliability-Gated Gaussian Evidence Fusion for Training-Free Test-Time Adaptation of Vision-Language Models}

\addauthor{Pedram MohajerAnsari\textsuperscript{\textbf{*}}}
{pmohaje@clemson.edu}{1}

\addauthor{Amir Salarpour\textsuperscript{\textbf{*}}}
{asalarp@clemson.edu}{1}

\addauthor{Run Wang}
{runw@clemson.edu}{1}

\addauthor{Mert D. Pesé}
{mpese@clemson.edu}{1}

\addinstitution{
Clemson University\\
Clemson, SC, USA
}

\runninghead{MohajerAnsari et al.}{GATE: Reliability-Gated Gaussian Evidence Fusion}

\makeatletter
\AtBeginDocument{%
  \def\hyper@natlinkstart#1{%
    \pdfstartlink
      attr{/Border [0 0 0]}
      goto name{cite.#1}\relax
  }%
  \def\hyper@natlinkend{%
    \pdfendlink
  }%
}
\makeatother

\begin{document}

\maketitle

\begingroup
\renewcommand{\thefootnote}{*}
\footnotetext{The first two authors contributed equally.}
\endgroup


\begin{abstract}
Vision-language models such as CLIP and SigLIP provide strong zero-shot recognition, but their predictions can degrade when deployed on target data that differ from the pretraining distribution. Test-time adaptation offers a practical way to improve robustness without source data or target labels, yet existing methods often rely on either prompt-side adaptation or image-side target evidence alone. In this work, we introduce \texttt{GATE}, a training-free two-pass transductive test-time adaptation framework that uses the unlabeled target set while keeping the image encoder, text encoder, and prompt parameters fully frozen. Instead of representing each class with a single prototype, \texttt{GATE} builds two complementary Gaussian sources of evidence in the shared vision-language feature space: a text Gaussian estimated from multiple language descriptions and an image Gaussian estimated from reliable unlabeled target samples. A class-wise reliability gate controls the influence of image-derived pseudo-evidence, and a score-level generalized Product-of-Experts fusion produces a normalized residual correction to the original zero-shot logits. Across fine-grained recognition datasets, ImageNet-family distribution shifts, multiple CLIP backbones, and SigLIP-B/16, \texttt{GATE} achieves the best average accuracy in every benchmark/backbone group. It improves zero-shot performance by an average of 5.41 points and outperforms the strongest non-\texttt{GATE} baseline by 1.94 points, demonstrating the benefit of reliability-gated distributional evidence for frozen VLM adaptation. Code is available at: \url{https://github.com/pedram-mohajer/GATE}
\end{abstract}

\section{Introduction}

Vision-language models (VLMs), such as CLIP~\cite{radford2021learning} and SigLIP~\cite{zhai2023sigmoid}, have become strong zero-shot recognition models by aligning visual and textual representations through large-scale pretraining. Their ability to classify images from natural-language descriptions makes them attractive for deployment without task-specific training data~\cite{lin2022frozen,guzhov2022audioclip,liu2023clip}. However, zero-shot predictions can degrade when test samples deviate from the pretraining distribution, as in domain shifts, style changes, sensor variations, or image corruptions~\cite{li2021align,li2023blip,zeng2023x}. Test-time adaptation (TTA) addresses this issue by adapting model behavior at deployment using unlabeled target data, without access to source training samples or target labels~\cite{ma2023swapprompt,shu2022tpt,karmanov2024efficient}.

Existing VLM adaptation methods mainly operate from either the language side or the image side. Language-side methods improve the text classifier by aggregating templates, selecting or reweighting prompts, or adapting prompt representations during inference~\cite{shu2022tpt,feng2023diverse,dong2026carprt}. These methods improve semantic coverage, but usually represent the target visual distribution only indirectly. Image-side methods instead exploit unlabeled target samples through pseudo-labels, augmentations, caches, retrieval, or distribution correction~\cite{farina2024frustratingly,karmanov2024efficient,zhou2023test,fan2025test,lee2025ra}. These methods can capture target-domain regularities, but their pseudo-evidence can be noisy and uneven across classes. As a result, existing approaches rarely provide a unified class-wise representation that models both prompt-induced semantic variability and target-domain visual variability while controlling the reliability of target evidence.

We introduce \texttt{GATE}, \textbf{G}aussian-guided \textbf{A}daptation with reliability-ga\textbf{T}ed \textbf{E}vidence fusion, a training-free test-time adaptation framework for frozen CLIP and SigLIP models. \texttt{GATE} moves beyond point estimates of class prototypes by representing each class with two complementary Gaussian sources in the shared vision-language feature space. A text Gaussian, estimated from multiple language descriptions, captures prompt-induced semantic variability and provides a stable class prior. An image Gaussian, estimated from reliable unlabeled target samples, summarizes target-domain visual structure. A class-wise reliability gate controls how strongly image-derived pseudo-evidence contributes to each class, and score-level generalized Product-of-Experts fusion produces a normalized residual correction to the original zero-shot logits. The image encoder, text encoder, and prompt parameters remain fixed throughout adaptation. \texttt{GATE} first estimates the image Gaussian from reliable unlabeled target samples and then uses the resulting Gaussian evidence to produce the normalized residual correction to the original zero-shot logits.

Experiments across fine-grained recognition datasets, ImageNet-family distribution shifts, and multiple VLM backbones show that \texttt{GATE} consistently improves zero-shot robustness without model updates. The gains come from complementing, rather than replacing, the original VLM classifier with reliability-aware evidence from both language descriptions and unlabeled target samples. The main contributions of this work are summarized as follows:
\begin{itemize}
\item We propose \texttt{GATE}, a training-free test-time adaptation framework for frozen CLIP and SigLIP models that uses only unlabeled target data and does not update image encoders, text encoders, or prompt parameters.

\item We introduce a dual Gaussian class representation in the shared vision-language feature space, where text Gaussians summarize prompt-induced semantic variability and image Gaussians summarize target-domain visual variability from reliable pseudo-evidence.

\item We develop a class-wise reliability-gated score-level fusion mechanism that combines text-derived and image-derived Gaussian evidence while limiting the influence of unreliable target statistics.

\item We conduct comprehensive experiments and ablations across multiple recognition datasets, distribution shifts, and VLM backbones, showing that \texttt{GATE} achieves strong and consistent improvements over zero-shot inference and representative TTA baselines.

\end{itemize}

\section{Related Work}
\label{sec:related}

\paragraph{Test-time adaptation for vision-language models.}
Test-time adaptation (TTA) aims to improve model predictions at deployment using unlabeled target data, without access to source training samples or target labels. For vision-language models, one line of work adapts the language side of the classifier. Test-time prompt tuning (TPT)~\cite{shu2022tpt} optimizes prompt parameters for each test sample using augmented views and entropy minimization, while related prompt-based methods improve robustness through stronger prompt aggregation, prompt selection, or class-aware prompt reweighting. For example, CARPRT~\cite{dong2026carprt} estimates class-specific prompt weights from unlabeled inference data, improving the text classifier without requiring gradients or access to model internals. These methods are effective when prompt wording and class descriptions are the main bottleneck, but they usually represent the target visual distribution only indirectly.

A second line of work adapts predictions from the image side. ZERO~\cite{farina2024frustratingly} shows that optimization-free confidence aggregation over augmented views can be a strong alternative to test-time prompt optimization. Distribution Normalization (DN)~\cite{zhou2023test} estimates a global target-domain representation shift and corrects image--text similarities without gradient updates. TDA~\cite{karmanov2024efficient} builds positive and negative caches from unlabeled target samples and combines cache-based scores with zero-shot logits. More recently, TT-RAA~\cite{fan2025test} uses streaming Gaussian target statistics and retrieval augmentation to inject target-domain evidence into CLIP predictions. These methods capture useful visual regularities, but the adaptation signal is often represented through global statistics, instance-level memories, or retrieval databases, and its reliability can vary substantially across classes. Although TT-RAA also uses Gaussian target statistics, it uses them for retrieval-augmented image-side adaptation, whereas \texttt{GATE} combines text-side and image-side Gaussian evidence through class-wise reliability-gated score fusion.

Optimization-based methods also explore richer use of target samples. BITTA~\cite{sun2026bilateral} uses low-entropy samples for learning representative target-domain features and high-entropy samples for unlearning unreliable patterns. While effective, such methods update model components during test-time adaptation and therefore differ from our fully frozen setting.

\paragraph{Positioning of \texttt{GATE}.}
Existing VLM adaptation methods improve zero-shot robustness through prompt optimization, augmentation-based voting, global distribution correction, dynamic caches, prompt reweighting, or retrieval-augmented prediction. In contrast, \texttt{GATE} models both sides of the VLM classifier as complementary class-wise evidence in the shared embedding space. A text Gaussian captures prompt-induced semantic variability from language descriptions, while an image Gaussian summarizes target-domain visual structure from reliable unlabeled samples. A class-wise reliability gate controls the influence of image-derived pseudo-evidence before score-level fusion with the original zero-shot logits. This keeps the image encoder, text encoder, and prompt parameters fully frozen while combining semantic language priors with target-domain visual statistics.

\section{Method}
\label{sec:method}

We introduce \texttt{GATE}, a training-free test-time adaptation method
for frozen vision-language models under distribution shift. We consider
a source-free and label-free setting in which only unlabeled target
images are available at deployment.

\begin{figure*}[t]
\centering
\includegraphics[width=1\linewidth]{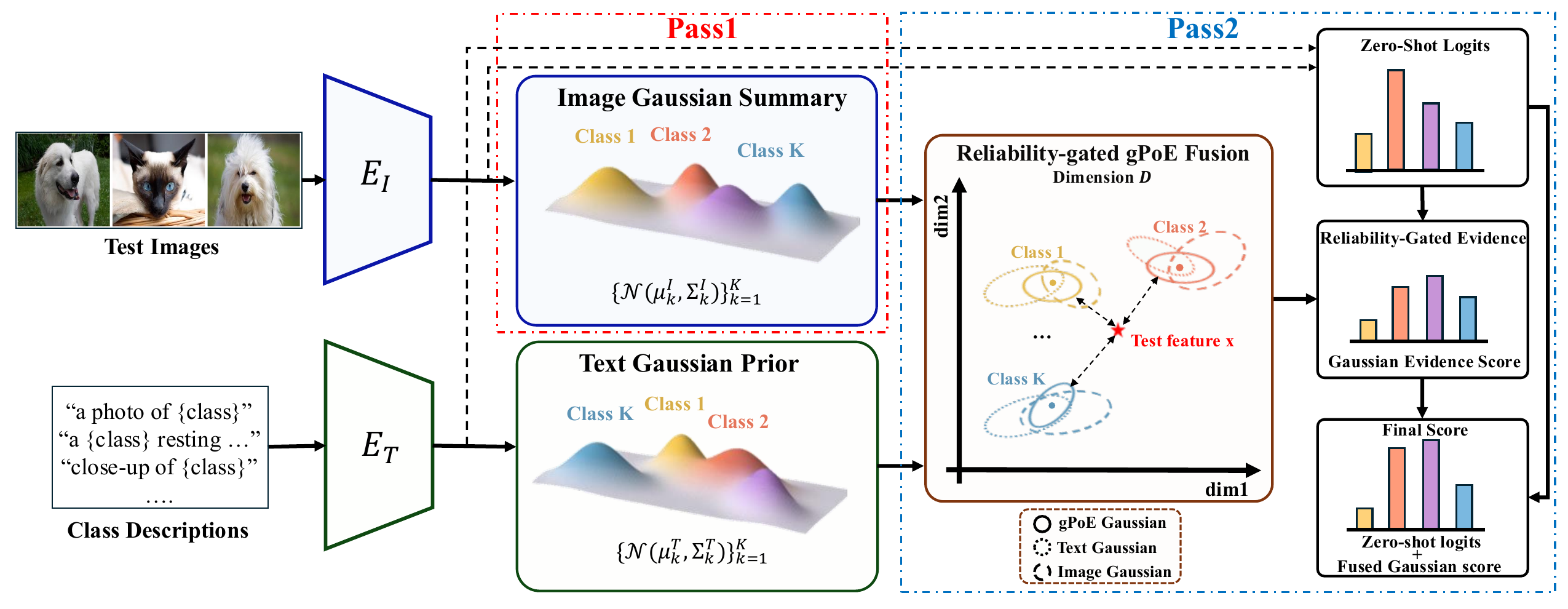}
\caption{
Overview of \texttt{GATE}. A frozen VLM first produces image features and zero-shot logits for unlabeled target images. In Pass~1, \texttt{GATE} uses the unlabeled target split to select
reliable pseudo-evidence and estimate class-specific image means together with a shared pooled covariance, while the text Gaussian prior is computed from language descriptions and kept fixed. In Pass~2, the text and image Gaussian statistics are converted into compatibility scores and combined through class-wise reliability-gated score-level generalized Product-of-Experts fusion. The resulting normalized Gaussian evidence is added as a residual correction to the original zero-shot logits to obtain the final prediction. The image encoder, text encoder, and prompt parameters remain frozen throughout, and no target labels are used.
}
\label{fig:method}
\end{figure*}

The image encoder, text encoder, and prompt parameters remain fixed.
\texttt{GATE} adapts the zero-shot classifier by estimating class-wise
distributional evidence in the shared vision-language embedding space.

As shown in Figure~\ref{fig:method}, each class is represented by two complementary Gaussian sources. A text Gaussian is estimated from multiple language descriptions and provides a stable semantic prior. An image Gaussian is estimated from reliable pseudo-evidence in the unlabeled target set and captures target-domain visual structure. A class-wise reliability gate controls the influence of image-derived evidence, and score-level generalized Product-of-Experts fusion produces a calibrated residual correction to the original zero-shot logits. We focus on the two-pass transductive setting: target images are first forwarded through the frozen VLM to estimate pseudo-evidence and class statistics, and final predictions are then obtained by reliability-gated Gaussian evidence fusion. 

We use these Gaussians as local distributional summaries in the normalized VLM embedding space, rather than as a full generative model of image or text features. The two sources play complementary roles: the text Gaussian provides class-specific semantic structure that is fixed across the target set, while the image Gaussian incorporates visual structure observed in the current target distribution. Their combination allows \texttt{GATE} to preserve the original semantic classifier while introducing target-specific evidence only when such evidence is sufficiently reliable.

\begin{table*}[b]
\centering
\scriptsize
\setlength{\tabcolsep}{3.0pt}
\renewcommand{\arraystretch}{1.08}
\caption{
Summary of clean test-time adaptation results across benchmark groups and frozen VLM backbones. Each row reports the average top-1 accuracy over datasets in the corresponding benchmark group. The best baseline excludes \texttt{GATE}. Gains are reported relative to zero-shot inference and to the strongest non-\texttt{GATE} baseline.
}
\label{tab:main_summary}
\begin{adjustbox}{max width=\textwidth}
\begin{tabular}{llccccc}
\toprule
Backbone & Benchmark & Zero-shot & Best baseline & \textsc{GATE} & Gain vs. Zero-shot & Gain vs. Best \\
\midrule
SigLIP-B/16 & Fine-grained & 73.34 & CARPRT (75.40) & \textbf{77.80} & +4.46 & +2.40 \\
\rowcolor{rowgray} CLIP ViT-B/16 & Fine-grained & 64.82 & CARPRT (67.35) & \textbf{69.25} & +4.43 & +1.90 \\
CLIP ViT-B/16 & ImageNet-family & 58.57 & Zero (63.65) & \textbf{65.66} & +7.09 & +2.01 \\
\rowcolor{rowgray} CLIP ViT-B/32 & Fine-grained & 60.57 & CARPRT (62.56) & \textbf{64.87} & +4.30 & +2.31 \\
CLIP ViT-B/32 & ImageNet-family & 49.89 & Zero (55.34) & \textbf{57.08} & +7.19 & +1.74 \\
\rowcolor{rowgray} CLIP ViT-L/14 & Fine-grained & 70.99 & CARPRT (74.08) & \textbf{75.47} & +4.48 & +1.39 \\
CLIP ViT-L/14 & ImageNet-family & 70.92 & Zero (74.98) & \textbf{76.84} & +5.92 & +1.86 \\
\midrule
Average & -- & -- & -- & -- & +5.41 & +1.94 \\
\bottomrule
\end{tabular}
\end{adjustbox}
\vspace{2pt}
\begin{flushleft}
\footnotesize
\textit{Note.} Values are average top-1 accuracies over datasets within each benchmark group. The last two columns report absolute percentage-point gains.
\end{flushleft}
\vspace{-15pt}
\end{table*}

\subsection{Frozen Zero-Shot VLM Classifier}

Let $E_I(\cdot)$ and $E_T(\cdot)$ denote the image and text encoders of a frozen VLM. For a $K$-class task with class names $\{c_k\}_{k=1}^{K}$, the standard zero-shot classifier represents each class by a normalized text prototype $t_k \in \mathbb{R}^{D}$, obtained by encoding and averaging prompt-based text embeddings. For a target image $x_i$, the normalized image feature is
\begin{equation}
    f_i =
    \frac{E_I(x_i)}{\|E_I(x_i)\|_2}.
\end{equation}
The zero-shot logit for class $k$ is
\begin{equation}
    z^{\mathrm{zs}}_{ik}
    =
    \tau \langle f_i,t_k\rangle ,
\end{equation}
where $\tau$ is the fixed logit scale of the pretrained VLM and $\langle \cdot,\cdot\rangle$ denotes cosine similarity. The zero-shot prediction is
\begin{equation}
    \hat{y}_i^{\mathrm{zs}}
    =
    \arg\max_{k} z^{\mathrm{zs}}_{ik}.
\end{equation}

\texttt{GATE} keeps the zero-shot classifier unchanged and applies a reliability-gated Gaussian evidence correction to $z^{\mathrm{zs}}_{ik}$.


\subsection{Text Gaussian Prior from Language Descriptions}

A standard zero-shot classifier compresses the language information of each class into a single text prototype. \texttt{GATE} instead preserves prompt-induced variation by constructing a text Gaussian prior for each class.  

For class $k$, let $\mathcal{P}_k=\{p_{k,1},\dots,p_{k,N_k}\}$ denote its prompt templates or language descriptions.
Each description is encoded and normalized as

\begin{equation}
    u_{k,j}
    =
    \frac{E_T(p_{k,j})}{\|E_T(p_{k,j})\|_2},
\end{equation}

\noindent The text Gaussian is
$\mathcal{Q}^{T}_k=\mathcal{N}(m^{T}_k,C^{T}_k)$,
where
$m^{T}_k=\frac{1}{N_k}\sum_{j=1}^{N_k}u_{k,j}$
and
\begin{equation}
    C^{T}_k
    =
    \frac{1}{N_k-1}
    \sum_{j=1}^{N_k}
    (u_{k,j}-m^{T}_k)
    (u_{k,j}-m^{T}_k)^\top
    +
    \lambda_T I,
    \qquad
    \lambda_T=0.01.
    \label{eq:text_covariance}
\end{equation}

The mean captures the class-level semantic center, while the covariance captures variation across descriptions. Unlike the standard zero-shot prototype, which averages this variation into a single vector, the Gaussian representation retains second-order information about how the language descriptions vary around the class center.

Since the number of descriptions is typically much smaller than the
embedding dimension, directly inverting the covariance can be unstable.
We therefore use a class-specific rank-$r_T$ truncated precision with
$r_T=15$. Let
$C^{T}_k=U_k\Lambda_kU_k^\top$
denote the eigendecomposition of the text covariance, and let
$U_{k,r_T}$ and $\Lambda_{k,r_T}$ contain the retained eigenvectors
and eigenvalues. Before inversion, we define

\[
    \widetilde{\Lambda}_{k,r_T}
    =
    \operatorname{diag}
    \left(
    \max\{\lambda_{k,1},\lambda_{\mathrm{floor}}\},
    \ldots,
    \max\{\lambda_{k,r_T},\lambda_{\mathrm{floor}}\}
    \right),
\]
where $\lambda_{\mathrm{floor}}>0$ is a fixed numerical floor.

The truncated text precision is then
\begin{equation}
    \Pi^{T}_k
    =
    U_{k,r_T}
    \widetilde{\Lambda}_{k,r_T}^{-1}
    U_{k,r_T}^{\top}.
    \label{eq:text_precision}
\end{equation}
The class-specific text precisions are computed once, cached, and kept
fixed throughout adaptation.


\subsection{Image Gaussian from Reliable Target Evidence}

The text Gaussian provides a stable semantic prior, but it does not capture target-domain visual structure. In Pass~1 of the two-pass transductive protocol, \texttt{GATE} uses the full unlabeled target split to construct image-side statistics from reliable pseudo-evidence produced by the frozen zero-shot classifier. Given the zero-shot logits for image $x_i$, we compute sparse class responsibilities over its top-5 predicted classes:
\begin{equation}
    \pi_{ik}
    =
    \frac{
    \mathbf{1}
    \!\left[k\in\mathcal{T}_{q_p}(i)\right]
    \exp\!\left(z^{\mathrm{zs}}_{ik}/\tau_p\right)
    }{
    \displaystyle
    \sum_{\ell\in\mathcal{T}_{q_p}(i)}
    \exp\!\left(z^{\mathrm{zs}}_{i\ell}/\tau_p\right)
    },
    \qquad
    q_p=\min(5,K),
    \quad
    \tau_p=0.5.
    \label{eq:responsibilities}
\end{equation}
For each class $k$, we retain at most $B=20$ target samples with the
largest responsibilities $\pi_{ik}$ and denote the resulting evidence
set by $\mathcal{E}_k$. Unless otherwise stated, all denominator and standard-deviation stabilizers use a fixed $\epsilon>0$.
Using the responsibilities as weights, we estimate the class-specific image mean as
\begin{equation}
    m^{I}_k =
    \frac{
    \sum_{i \in \mathcal{E}_k}\pi_{ik}f_i
    }
    {
    \sum_{i \in \mathcal{E}_k}\pi_{ik}+\epsilon
    } .
\end{equation}
The selected evidence also provides a weighted covariance estimate for each class. Rather than maintaining and inverting a separate image covariance for every class, these covariance estimates are pooled into a single empirical covariance $\widehat{C}^{I}$. We then apply fixed shrinkage and ridge regularization:
\begin{equation}
    C^{I}
    =
    (1-\rho)\widehat{C}^{I}
    +
    \rho\,
    \frac{\operatorname{tr}(\widehat{C}^{I})}{D}I
    +
    \lambda_I I,
    \qquad
    \rho=0.5,
    \quad
    \lambda_I=0.01.
    \label{eq:image_covariance}
\end{equation}
The image-side representation is therefore $\mathcal{Q}^{I}_k=\mathcal{N}(m^{I}_k,C^{I})$, with a class-specific mean and a covariance shared across classes. This design preserves class-specific target centers while avoiding a separate high-dimensional covariance inversion for every class. 
The corresponding image precision is computed once from the pooled
covariance.

During Pass~2, it is reused for all classes. We use this Gaussian only
as a regularized second-order compatibility summary of target evidence,
rather than as a full generative model.

We estimate the reliability of each class from its selected pseudo-evidence as
\begin{equation}
    n^{\mathrm{eff}}_k
    =
    \sum_{i\in\mathcal{E}_k}\pi_{ik},
    \qquad
    r_k
    =
    \frac{1}{|\mathcal{E}_k|+\epsilon}
    \sum_{i\in\mathcal{E}_k}\pi_{ik}.
\end{equation}
Here, $n^{\mathrm{eff}}_k$ measures the amount of available evidence, while $r_k$ measures its average confidence. These quantities are used by the reliability gate to down-weight classes with weak or ambiguous target evidence.



\begin{table*}[t]
\centering
\footnotesize
\setlength{\tabcolsep}{3.0pt}
\renewcommand{\arraystretch}{1.08}
\caption{
ImageNet-family distribution-shift results using CLIP ViT-B/16. Dataset rows report top-1 accuracy as mean$\pm$std over three random seeds; the Mean row reports the average across datasets.
}
\label{tab:vit_b16_imagenet_family}
\begin{adjustbox}{max width=\textwidth}
\begin{tabular}{l c| cccccccc}
\toprule
Dataset & Zero-shot & Zero & TPT & BITTA & TT-RAA & CARPRT & DN & TDA & \textsc{GATE} (ours) \\
\midrule
ImageNet & 66.63$\pm$0.00 & \underline{69.29$\pm$0.02} & 69.08$\pm$0.06 & 67.43$\pm$0.25 & 67.36$\pm$0.04 & 68.83$\pm$0.00 & 66.94$\pm$0.04 & 68.28$\pm$0.08 & \cellcolor{oursgreen} \textbf{70.92$\pm$0.09} \\
\rowcolor{rowgray} ImageNetA & 47.85$\pm$0.00 & \underline{59.80$\pm$0.19} & 57.68$\pm$0.35 & 52.25$\pm$0.79 & 48.76$\pm$0.25 & 51.60$\pm$0.00 & 48.72$\pm$0.05 & 49.39$\pm$0.11 & \cellcolor{oursgreen} \textbf{60.34$\pm$0.31} \\
ImageNetR & 73.84$\pm$0.00 & 77.21$\pm$0.12 & 76.16$\pm$0.13 & \underline{78.51$\pm$0.42} & 74.44$\pm$0.06 & 77.56$\pm$0.00 & 75.56$\pm$0.06 & 75.37$\pm$0.01 & \cellcolor{oursgreen} \textbf{80.42$\pm$0.17} \\
\rowcolor{rowgray} ImageNetSketch & 45.95$\pm$0.00 & 48.31$\pm$0.01 & 47.13$\pm$0.17 & 47.69$\pm$0.09 & 47.65$\pm$0.09 & 46.89$\pm$0.00 & 47.05$\pm$0.09 & \underline{48.66$\pm$0.08} & \cellcolor{oursgreen} \textbf{50.95$\pm$0.03} \\
\midrule
\textbf{Mean} & 58.57 & \underline{63.65} & 62.51 & 61.47 & 59.55 & 61.22 & 59.57 & 60.43 & \cellcolor{oursgreen} \textbf{65.66} \\
\bottomrule
\end{tabular}
\end{adjustbox}

\vspace{2pt}
\begin{flushleft}
\footnotesize
\textit{Note.} Bold and underline indicate the best and second-best
methods, respectively.
\end{flushleft}
\vspace{-15pt}
\end{table*}


\begin{table*}[b]
\centering
\footnotesize
\setlength{\tabcolsep}{3.0pt}
\renewcommand{\arraystretch}{1.08}
\caption{
ImageNet-family distribution-shift results using CLIP ViT-B/32. Dataset rows report top-1 accuracy as mean$\pm$std over three random seeds; the Mean row reports the average across datasets.
}
\label{tab:vit_b32_imagenet_family}
\begin{adjustbox}{max width=\textwidth}
\begin{tabular}{l c| cccccccc}
\toprule
Dataset & Zero-shot & Zero & TPT & BITTA & TT-RAA & CARPRT & DN & TDA & \textsc{GATE} (ours) \\
\midrule
ImageNet & 61.88$\pm$0.00 & \underline{65.11$\pm$0.04} & 64.15$\pm$0.08 & 62.65$\pm$0.18 & 62.44$\pm$0.07 & 63.65$\pm$0.00 & 62.12$\pm$0.05 & 63.38$\pm$0.08 & \cellcolor{oursgreen} \textbf{66.60$\pm$0.12} \\
\rowcolor{rowgray} ImageNetA & 30.03$\pm$0.00 & \underline{41.13$\pm$0.10} & 37.46$\pm$0.07 & 33.23$\pm$0.56 & 30.68$\pm$0.18 & 32.24$\pm$0.00 & 30.36$\pm$0.07 & 31.32$\pm$0.21 & \cellcolor{oursgreen} \textbf{41.88$\pm$0.30} \\
ImageNetR & 66.84$\pm$0.00 & \underline{71.56$\pm$0.13} & 69.46$\pm$0.09 & 70.53$\pm$0.26 & 67.09$\pm$0.02 & 69.01$\pm$0.00 & 67.94$\pm$0.03 & 68.00$\pm$0.04 & \cellcolor{oursgreen} \textbf{73.98$\pm$0.14} \\
\rowcolor{rowgray} ImageNetSketch & 40.80$\pm$0.01 & \underline{43.57$\pm$0.07} & 41.17$\pm$0.08 & 39.09$\pm$1.30 & 41.75$\pm$0.17 & 40.85$\pm$0.00 & 41.25$\pm$0.07 & 43.09$\pm$0.06 & \cellcolor{oursgreen} \textbf{45.87$\pm$0.07} \\
\midrule
\textbf{Mean} & 49.89 & \underline{55.34} & 53.06 & 51.37 & 50.49 & 51.44 & 50.42 & 51.45 & \cellcolor{oursgreen} \textbf{57.08} \\
\bottomrule
\end{tabular}
\end{adjustbox}

\vspace{2pt}
\begin{flushleft}
\footnotesize
\textit{Note.} Bold and underline indicate the best and second-best
methods, respectively.
\end{flushleft}
\vspace{-15pt}
\end{table*}

\subsection{Precision-Weighted Gaussian Evidence}

We convert the text and image statistics into class-wise second-order compatibility scores. The text precision $\Pi^{T}_k$ is class-specific and fixed from the language descriptions, whereas the image precision
$\Pi^{I}=(C^{I})^{-1}$ is shared across classes and computed once from the pooled target covariance.

For target feature $f_i$, the corresponding precision-weighted energies are
\begin{equation}
\begin{aligned}
    d^{T}_{ik}
    &=
    (f_i-m^{T}_k)^\top
    \Pi^{T}_k
    (f_i-m^{T}_k), \\
    d^{I}_{ik}
    &=
    (f_i-m^{I}_k)^\top
    \Pi^{I}
    (f_i-m^{I}_k).
\end{aligned}
\end{equation}
These Mahalanobis-type energies penalize deviations more strongly along low-uncertainty directions. We use them as regularized second-order compatibility measures rather than as generative likelihoods.

To make the two evidence sources comparable across classes, each energy vector is standardized independently for each target sample:
\begin{equation}
    h^{e}_{ik}
    =
    -\,
    \frac{
    d^{e}_{ik}-\mu^{e}_{i}
    }{
    \sigma^{e}_{i}+\epsilon
    },
    \qquad e\in\{T,I\},
\end{equation}
where $\mu^{e}_{i}$ and $\sigma^{e}_{i}$ are respectively the mean and standard deviation of
$\{d^{e}_{i\ell}\}_{\ell=1}^{K}$ over classes. Thus, larger $h^{e}_{ik}$ indicates stronger compatibility between feature $f_i$ and class $k$ under evidence source $e$.


\subsection{Reliability-Gated Score-Level gPoE}

The text Gaussian provides a stable semantic prior, while the image Gaussian captures target-domain structure but depends on potentially noisy pseudo-evidence. We therefore control the image contribution using a class-wise reliability gate:
\begin{equation}
    \omega_k
    =
    \operatorname{clip}
    \left(
    \left(
    \frac{n^{\mathrm{eff}}_k}
    {n^{\mathrm{eff}}_k+\kappa}
    \right)^{\gamma}
    r_k^{\delta},
    0,
    \omega_{\max}
    \right).
    \label{eq:reliability_gate}
\end{equation}
The evidence-count factor increases with the amount of selected target
evidence, whereas $r_k$ reflects its average confidence. We use
$\kappa=20$, $\gamma=\delta=1$, and $\omega_{\max}=0.5$ for all datasets
and backbones; the exponent sweeps in Sec.~5 are diagnostic sensitivity
analyses only. The clipping bound limits image-derived evidence when
pseudo-evidence is weak or ambiguous. Given the normalized text and image
evidence scores $h^{T}_{ik}$ and $h^{I}_{ik}$, \texttt{GATE} performs
score-level generalized Product-of-Experts fusion as
\begin{equation}
    h^{\mathrm{gpoe}}_{ik}
    =
    (1-\omega_k)h^{T}_{ik}
    +
    \omega_k h^{I}_{ik}.
    \label{eq:gpoe}
\end{equation}
Thus, weak or unreliable target evidence keeps the fused score close to the text prior, whereas reliable target evidence shifts it toward the target-domain image evidence.


\subsection{Final Prediction}

Because the fused Gaussian evidence and zero-shot logits can have different scales, we calibrate the evidence independently for each target sample. Let
\begin{equation}
    \bar{h}^{\mathrm{gpoe}}_{ik}
    =
    \frac{
    h^{\mathrm{gpoe}}_{ik}
    -
    \mathrm{mean}_{\ell}
    (h^{\mathrm{gpoe}}_{i\ell})
    }
    {
    \mathrm{std}_{\ell}
    (h^{\mathrm{gpoe}}_{i\ell})
    +\epsilon
    } .
\end{equation}
The residual correction is restricted to the top-$q_r$ classes under
the frozen zero-shot classifier. Here, $\alpha$ is the residual strength,
$s_r$ is a fixed calibration scale, and $c$ is the clipping bound.
We use $\alpha=0.10$, $s_r=1.5$, $q_r=\min(15,K)$, and $c=4$ throughout
all experiments.

\begin{equation}
    \widetilde{h}^{\mathrm{gpoe}}_{ik}
    =
    \frac{\alpha}{s_r}\,
    \mathbf{1}
    \!\left[k\in\mathcal{T}_{q_r}(i)\right]
    \operatorname{clip}
    \left(
    \bar{h}^{\mathrm{gpoe}}_{ik},
    -c,
    c
    \right).
    \label{eq:residual_correction}
\end{equation}

The final score and prediction are
\begin{equation}
    s_{ik}
    =
    z^{\mathrm{zs}}_{ik}
    +
    \widetilde{h}^{\mathrm{gpoe}}_{ik},
    \qquad
    \hat{y}_i
    =
    \arg\max_k s_{ik}.
    \label{eq:final_prediction}
\end{equation}

Thus, \texttt{GATE} preserves the frozen zero-shot classifier and adds
only a normalized, bounded, reliability-gated Gaussian evidence
correction.


\section{Experiments}
\label{sec:experiments}

We evaluate \texttt{GATE} on clean test-time adaptation for vision-language recognition under distribution shift, using only unlabeled target data without access to source training samples or target labels. We consider diverse fine-grained and domain-specific recognition datasets together with ImageNet-family distribution shifts, and evaluate across multiple CLIP backbones and SigLIP-B/16. We compare against representative VLM test-time adaptation methods to assess whether reliability-gated Gaussian evidence provides consistent improvements over zero-shot inference across datasets, distribution shifts, and VLM backbones.

\subsection{Experimental Setup}


\noindent\textbf{Datasets.}
We evaluate on two benchmark groups. The first group contains diverse fine-grained and domain-specific recognition datasets: 
Caltech101~\cite{fei2004learning},
DTD~\cite{cimpoi2014describing},
EuroSAT~\cite{helber2019eurosat},
FGVC-Aircraft~\cite{maji2013fine},
Flowers102~\cite{nilsback2008automated},
Food101~\cite{bossard2014food},
OxfordPets~\cite{parkhi2012cats},
StanfordCars~\cite{krause2013object},
SUN397~\cite{xiao2010sun}, and
UCF101~\cite{soomro2012ucf101}.
The second group is the ImageNet-family distribution-shift benchmark,
consisting of ImageNet~\cite{deng2009imagenet},
ImageNet-A~\cite{hendrycks2021natural},
ImageNet-R~\cite{hendrycks2021many}, and
ImageNet-Sketch~\cite{wang2019learning}.

\noindent\textbf{Backbones.}
We evaluate CLIP ViT-B/32, ViT-B/16, and ViT-L/14, together with SigLIP-B/16, covering multiple model scales and VLM variants. This evaluation allows us to examine whether the proposed Gaussian evidence fusion remains effective across different representation capacities and pretraining formulations. For \textsc{GATE}, the image encoder, text encoder, and prompt parameters
remain fixed throughout adaptation.


\begin{table*}[t]
\centering
\footnotesize
\setlength{\tabcolsep}{3.0pt}
\renewcommand{\arraystretch}{1.08}
\caption{
Fine-grained and diverse recognition results using CLIP ViT-B/16. Dataset rows report top-1 accuracy as mean$\pm$std over three random seeds; the Mean row reports the average across datasets.
}
\label{tab:vit_b16_fine_grained}
\begin{adjustbox}{max width=\textwidth}
\begin{tabular}{l c| cccccccc}
\toprule
Dataset & Zero-shot & Zero & TPT & BITTA & TT-RAA & CARPRT & DN & TDA & \textsc{GATE} (ours) \\
\midrule
Caltech101 & 91.97$\pm$0.00 & 92.86$\pm$0.25 & 93.41$\pm$0.14 & 92.13$\pm$0.74 & 92.94$\pm$0.15 & \underline{94.56$\pm$0.00} & 93.08$\pm$0.08 & 93.02$\pm$0.24 & \cellcolor{oursgreen} \textbf{94.85$\pm$0.00} \\
\rowcolor{rowgray} DTD & 44.62$\pm$0.00 & 44.52$\pm$0.28 & 45.74$\pm$0.45 & 44.62$\pm$0.64 & 46.63$\pm$0.43 & \textbf{48.88$\pm$0.00} & 45.59$\pm$0.18 & 46.69$\pm$0.52 & \cellcolor{oursgreen} \underline{47.64$\pm$0.00} \\
EuroSAT & 52.06$\pm$0.00 & 44.76$\pm$0.01 & 37.03$\pm$0.09 & 41.29$\pm$4.92 & 57.28$\pm$0.43 & 54.68$\pm$0.00 & 56.27$\pm$0.45 & \underline{60.14$\pm$0.57} & \cellcolor{oursgreen} \textbf{63.88$\pm$0.00} \\
\rowcolor{rowgray} Aircraft & 24.75$\pm$0.00 & \textbf{27.00$\pm$0.33} & 23.94$\pm$0.00 & 20.63$\pm$0.74 & 24.35$\pm$0.52 & 24.87$\pm$0.00 & 25.63$\pm$0.12 & 25.38$\pm$0.43 & \cellcolor{oursgreen} \underline{26.10$\pm$0.00} \\
Flowers102 & 70.89$\pm$0.00 & 70.82$\pm$0.08 & 68.85$\pm$0.16 & 69.73$\pm$0.33 & \underline{72.49$\pm$0.06} & 71.30$\pm$0.00 & 70.63$\pm$0.05 & 71.00$\pm$0.41 & \cellcolor{oursgreen} \textbf{73.24$\pm$0.00} \\
\rowcolor{rowgray} Food101 & 83.13$\pm$0.00 & 84.75$\pm$0.05 & \textbf{86.27$\pm$0.03} & 85.79$\pm$0.28 & 83.55$\pm$0.05 & 85.88$\pm$0.00 & 84.22$\pm$0.06 & 84.34$\pm$0.06 & \cellcolor{oursgreen} \underline{86.07$\pm$0.00} \\
OxfordPets & 87.63$\pm$0.00 & 87.53$\pm$0.04 & 87.10$\pm$0.06 & 89.32$\pm$0.40 & 88.46$\pm$0.18 & \underline{89.53$\pm$0.00} & 87.52$\pm$0.05 & 88.67$\pm$0.21 & \cellcolor{oursgreen} \textbf{90.98$\pm$0.00} \\
\rowcolor{rowgray} StanfordCars & 65.24$\pm$0.00 & \textbf{68.04$\pm$0.13} & 67.91$\pm$0.27 & 63.49$\pm$0.47 & 66.17$\pm$0.54 & 66.06$\pm$0.00 & 65.63$\pm$0.09 & 66.35$\pm$0.07 & \cellcolor{oursgreen} \underline{68.01$\pm$0.00} \\
SUN397 & 62.23$\pm$0.00 & 64.43$\pm$0.12 & 65.19$\pm$0.13 & 66.02$\pm$0.43 & 64.73$\pm$0.02 & \underline{66.92$\pm$0.00} & 63.43$\pm$0.10 & 65.20$\pm$0.12 & \cellcolor{oursgreen} \textbf{69.11$\pm$0.00} \\
\rowcolor{rowgray} UCF101 & 65.64$\pm$0.00 & 67.80$\pm$0.24 & 68.77$\pm$0.36 & 67.02$\pm$0.71 & 67.83$\pm$0.12 & \underline{70.82$\pm$0.00} & 65.81$\pm$0.16 & 67.94$\pm$0.29 & \cellcolor{oursgreen} \textbf{72.59$\pm$0.00} \\
\midrule
\textbf{Mean} & 64.82 & 65.25 & 64.42 & 64.00 & 66.44 & \underline{67.35} & 65.78 & 66.87 & \cellcolor{oursgreen} \textbf{69.25} \\
\bottomrule
\end{tabular}
\end{adjustbox}

\vspace{2pt}
\begin{flushleft}
\footnotesize
\textit{Note.} Bold and underline indicate the best and second-best
methods, respectively.
\end{flushleft}
\vspace{-15pt}
\end{table*}


\begin{table*}[b]
\centering
\footnotesize
\setlength{\tabcolsep}{3.0pt}
\renewcommand{\arraystretch}{1.08}
\caption{
Fine-grained and diverse recognition results using CLIP ViT-B/32. Dataset rows report top-1 accuracy as mean$\pm$std over three random seeds; the Mean row reports the average across datasets.
}
\label{tab:vit_b32_fine_grained}
\begin{adjustbox}{max width=\textwidth}
\begin{tabular}{l c| cccccccc}
\toprule
Dataset & Zero-shot & Zero & TPT & BITTA & TT-RAA & CARPRT & DN & TDA & \textsc{GATE} (ours) \\
\midrule
Caltech101 & 90.51$\pm$0.00 & 90.78$\pm$0.12 & 90.74$\pm$0.24 & 91.20$\pm$0.61 & 91.47$\pm$0.17 & \textbf{93.10$\pm$0.00} & 91.35$\pm$0.10 & 91.44$\pm$0.12 & \cellcolor{oursgreen} \underline{92.86$\pm$0.00} \\
\rowcolor{rowgray} DTD & 42.38$\pm$0.00 & 43.44$\pm$0.36 & \underline{44.66$\pm$0.33} & 42.59$\pm$0.35 & 43.01$\pm$0.22 & 42.32$\pm$0.00 & 43.93$\pm$0.15 & 43.56$\pm$0.20 & \cellcolor{oursgreen} \textbf{49.17$\pm$0.00} \\
EuroSAT & 44.78$\pm$0.00 & 37.83$\pm$0.05 & 27.58$\pm$0.17 & 13.19$\pm$1.34 & \textbf{51.73$\pm$2.82} & 47.80$\pm$0.00 & 46.20$\pm$0.16 & 48.93$\pm$1.01 & \cellcolor{oursgreen} \underline{51.73$\pm$0.00} \\
\rowcolor{rowgray} Aircraft & 19.26$\pm$0.00 & \textbf{20.83$\pm$0.17} & 19.19$\pm$0.27 & 16.09$\pm$0.83 & 19.27$\pm$0.25 & 19.17$\pm$0.00 & \underline{20.09$\pm$0.15} & 19.94$\pm$0.18 & \cellcolor{oursgreen} 19.83$\pm$0.00 \\
Flowers102 & 66.26$\pm$0.00 & 66.52$\pm$0.18 & 63.38$\pm$0.18 & 64.85$\pm$0.21 & 67.32$\pm$0.37 & 66.30$\pm$0.00 & 65.96$\pm$0.15 & \underline{67.94$\pm$0.41} & \cellcolor{oursgreen} \textbf{68.53$\pm$0.00} \\
\rowcolor{rowgray} Food101 & 76.96$\pm$0.00 & 79.08$\pm$0.03 & \textbf{80.87$\pm$0.08} & \underline{80.64$\pm$0.14} & 77.38$\pm$0.06 & 80.14$\pm$0.00 & 78.15$\pm$0.07 & 78.44$\pm$0.10 & \cellcolor{oursgreen} 80.63$\pm$0.00 \\
OxfordPets & 83.86$\pm$0.00 & 84.97$\pm$0.11 & 84.17$\pm$0.09 & 84.86$\pm$0.63 & 84.19$\pm$0.77 & \underline{86.86$\pm$0.00} & 82.86$\pm$0.11 & 83.41$\pm$0.09 & \cellcolor{oursgreen} \textbf{88.31$\pm$0.00} \\
\rowcolor{rowgray} StanfordCars & 60.20$\pm$0.00 & \underline{64.02$\pm$0.26} & \textbf{64.28$\pm$0.09} & 59.24$\pm$0.20 & 60.65$\pm$0.23 & 60.86$\pm$0.00 & 60.76$\pm$0.08 & 61.13$\pm$0.15 & \cellcolor{oursgreen} 61.91$\pm$0.00 \\
SUN397 & 60.31$\pm$0.00 & 62.20$\pm$0.11 & 63.87$\pm$0.13 & 63.82$\pm$0.28 & 62.28$\pm$0.14 & \underline{64.05$\pm$0.00} & 61.19$\pm$0.03 & 62.74$\pm$0.04 & \cellcolor{oursgreen} \textbf{67.17$\pm$0.00} \\
\rowcolor{rowgray} UCF101 & 61.17$\pm$0.00 & 63.63$\pm$0.25 & 63.98$\pm$0.24 & 62.03$\pm$0.29 & 64.16$\pm$0.62 & \underline{65.03$\pm$0.00} & 62.33$\pm$0.14 & 64.32$\pm$0.22 & \cellcolor{oursgreen} \textbf{68.54$\pm$0.00} \\
\midrule
\textbf{Mean} & 60.57 & 61.33 & 60.27 & 57.85 & 62.15 & \underline{62.56} & 61.28 & 62.18 & \cellcolor{oursgreen} \textbf{64.87} \\
\bottomrule
\end{tabular}
\end{adjustbox}

\vspace{2pt}
\begin{flushleft}
\footnotesize
\textit{Note.} Bold and underline indicate the best and second-best
methods, respectively.
\end{flushleft}
\vspace{-15pt}
\end{table*}


\begin{table*}[t]
\centering
\footnotesize
\setlength{\tabcolsep}{3.0pt}
\renewcommand{\arraystretch}{1.08}
\caption{
Fine-grained and diverse recognition results using CLIP ViT-L/14. Dataset rows report top-1 accuracy as mean$\pm$std over three random seeds; the Mean row reports the average across datasets.
}
\label{tab:vit_l14_fine_grained}
\begin{adjustbox}{max width=\textwidth}
\begin{tabular}{l c| cccccccc}
\toprule
Dataset & Zero-shot & Zero & TPT & BITTA & TT-RAA & CARPRT & DN & TDA & \textsc{GATE} (ours) \\
\midrule
Caltech101 & 93.47$\pm$0.00 & 94.14$\pm$0.06 & 95.25$\pm$0.21 & 92.86$\pm$0.49 & 93.55$\pm$0.21 & \underline{96.51$\pm$0.00} & 94.14$\pm$0.10 & 93.60$\pm$0.10 & \cellcolor{oursgreen} \textbf{97.44$\pm$0.00} \\
\rowcolor{rowgray} DTD & 52.78$\pm$0.00 & 53.86$\pm$0.17 & 54.18$\pm$0.33 & 53.88$\pm$0.24 & 54.02$\pm$0.36 & \textbf{57.68$\pm$0.00} & 53.51$\pm$0.09 & 54.06$\pm$0.43 & \cellcolor{oursgreen} \underline{57.45$\pm$0.00} \\
EuroSAT & 57.86$\pm$0.00 & 50.16$\pm$0.06 & 42.38$\pm$0.14 & 53.21$\pm$1.12 & 65.12$\pm$0.94 & 64.73$\pm$0.00 & 61.10$\pm$0.51 & \underline{66.27$\pm$0.54} & \cellcolor{oursgreen} \textbf{67.40$\pm$0.00} \\
\rowcolor{rowgray} Aircraft & 32.64$\pm$0.00 & \textbf{37.24$\pm$0.10} & 32.47$\pm$0.37 & 24.75$\pm$1.59 & 33.54$\pm$0.12 & 32.34$\pm$0.00 & 33.15$\pm$0.09 & 33.35$\pm$0.30 & \cellcolor{oursgreen} \underline{34.02$\pm$0.00} \\
Flowers102 & 79.01$\pm$0.00 & 78.86$\pm$0.02 & 76.13$\pm$0.12 & 76.75$\pm$0.82 & \underline{80.28$\pm$0.12} & 78.93$\pm$0.00 & 78.93$\pm$0.07 & 79.62$\pm$0.33 & \cellcolor{oursgreen} \textbf{81.12$\pm$0.00} \\
\rowcolor{rowgray} Food101 & 88.25$\pm$0.00 & 89.17$\pm$0.06 & 90.74$\pm$0.02 & 90.42$\pm$0.31 & 88.51$\pm$0.03 & \textbf{91.25$\pm$0.00} & 89.16$\pm$0.02 & 89.41$\pm$0.04 & \cellcolor{oursgreen} \underline{91.07$\pm$0.00} \\
OxfordPets & 92.86$\pm$0.00 & 92.86$\pm$0.14 & 93.69$\pm$0.21 & 91.71$\pm$0.17 & 93.57$\pm$0.10 & \underline{94.11$\pm$0.00} & 93.19$\pm$0.03 & 93.40$\pm$0.22 & \cellcolor{oursgreen} \textbf{94.28$\pm$0.00} \\
\rowcolor{rowgray} StanfordCars & 76.13$\pm$0.00 & 78.10$\pm$0.03 & \underline{78.11$\pm$0.22} & 75.38$\pm$0.10 & 76.62$\pm$0.25 & 78.03$\pm$0.00 & 76.73$\pm$0.17 & 76.94$\pm$0.21 & \cellcolor{oursgreen} \textbf{79.23$\pm$0.00} \\
SUN397 & 66.25$\pm$0.00 & 68.19$\pm$0.12 & 69.36$\pm$0.05 & 70.17$\pm$0.46 & 68.55$\pm$0.11 & \underline{70.75$\pm$0.00} & 67.78$\pm$0.03 & 68.67$\pm$0.24 & \cellcolor{oursgreen} \textbf{73.42$\pm$0.00} \\
\rowcolor{rowgray} UCF101 & 70.63$\pm$0.00 & 71.97$\pm$0.08 & 75.15$\pm$0.26 & 74.61$\pm$0.73 & 72.53$\pm$0.12 & \underline{76.45$\pm$0.00} & 70.84$\pm$0.07 & 72.61$\pm$0.23 & \cellcolor{oursgreen} \textbf{79.30$\pm$0.00} \\
\midrule
\textbf{Mean} & 70.99 & 71.46 & 70.75 & 70.38 & 72.63 & \underline{74.08} & 71.85 & 72.79 & \cellcolor{oursgreen} \textbf{75.47} \\
\bottomrule
\end{tabular}
\end{adjustbox}

\vspace{2pt}
\begin{flushleft}
\footnotesize
\textit{Note.} Bold and underline indicate the best and second-best
methods, respectively.
\end{flushleft}
\vspace{-15pt}
\end{table*}

\noindent\textbf{Baselines.}
We compare against zero-shot inference and several representative VLM test-time adaptation methods. Zero-shot uses the frozen VLM classifier without target adaptation. Zero performs optimization-free test-time adaptation through confidence-based aggregation over augmented predictions. TPT performs test-time prompt tuning using augmented views and entropy minimization. BITTA adapts the model using bilateral learning and unlearning from low- and high-entropy target samples. DN performs training-free distribution normalization using target-domain statistics. TDA uses dynamic positive and negative caches built from unlabeled target samples. TT-RAA performs retrieval-augmented adaptation using a streaming Gaussian database. CARPRT performs class-aware prompt reweighting from unlabeled inference data.


\noindent\textbf{Protocol and implementation.}
All methods are source-free and label-free, start from the same pretrained backbone, and use the same dataset split, class names, base prompt set where applicable, and evaluation metric. All methods are evaluated without target-label tuning. GATE follows a two-pass transductive protocol rather than a strict online single-sample setting: Pass~1 uses the full unlabeled target split to select pseudo-evidence and estimate Gaussian statistics, and Pass~2 predicts using the resulting evidence. Baselines follow their standard protocols; methods requiring target-set statistics or caches are allowed access to the same unlabeled target split. For a fixed split, prompt set, hyperparameters, sample order, and tie-breaking, \textsc{GATE} is deterministic and uses no stochastic optimization, augmentation sampling, prompt updates, or encoder updates.


\noindent\textbf{Computational cost.}
\textsc{GATE} avoids per-class image covariance inversions by pooling the
image covariances into a single shrinkage-regularized precision matrix,
which is inverted once per dataset and reused across classes. After the
precision matrices are cached, scoring requires only matrix--vector
products and inner products, with no gradients or model updates.

\begin{table}[b]
\centering
\small
\caption{Adaptation protocol of the compared methods.}
\label{tab:protocols}
\setlength{\tabcolsep}{1.8pt}
\renewcommand{\arraystretch}{1.05}
\begin{tabular}{lcccccccc}
\toprule
 & Zero & TPT & BITTA & TDA & TT-RAA & DN & CARPRT & GATE \\
\midrule
Protocol
& PS
& PS
& OL
& SC
& SC
& TR
& TR
& TR \\
\bottomrule
\end{tabular}

\vspace{1pt}
\footnotesize
PS: per-sample; OL: online; SC: streaming-cache; TR: transductive.
\end{table}



\noindent\textbf{Metrics.}
We report top-1 classification accuracy. Dataset rows report
mean$\pm$standard deviation over three random seeds. Because
\textsc{GATE} is deterministic for a fixed split, prompt set,
hyperparameters, sample order, and tie-breaking, most \textsc{GATE}
entries are $\pm 0.00$; the small nonzero deviations in the 1000-class
ImageNet-family experiments arise from rare pseudo-evidence ordering or
tie-breaking effects. The final Mean row reports a single average across
datasets. Bold and underline indicate the best and second-best methods,
respectively.

\subsection{Main Results}

Table~\ref{tab:main_summary} summarizes the main comparison across benchmark groups and VLM backbones. \textsc{GATE} achieves the best average accuracy in every setting. Across the seven benchmark/backbone groups, \textsc{GATE} improves the zero-shot classifier by an average of 5.41 percentage points and outperforms the strongest non-\textsc{GATE} baseline by an average of 1.94 points. Across 52 dataset-backbone settings, \textsc{GATE} obtains the best accuracy in 39 cases and ranks among the top two methods in 49 cases. The gains are particularly pronounced on the ImageNet-family benchmark, where \textsc{GATE} improves over zero-shot by +7.09 points for CLIP ViT-B/16 and +7.19 points for CLIP ViT-B/32.

\subsection{Fine-Grained and Diverse Recognition Results}


\begin{table*}[b]
\centering
\footnotesize
\setlength{\tabcolsep}{3.0pt}
\renewcommand{\arraystretch}{1.08}
\caption{
Fine-grained and diverse recognition results using SigLIP-B/16. Dataset rows report top-1 accuracy as mean$\pm$std over three random seeds; the Mean row reports the average across datasets.
}
\label{tab:siglip_b16_fine_grained}
\begin{adjustbox}{max width=\textwidth}
\begin{tabular}{l c| cccccccc}
\toprule
Dataset & Zero-shot & Zero & TPT & BITTA & TT-RAA & CARPRT & DN & TDA & \textsc{GATE} (ours) \\
\midrule
Caltech101 & 96.84$\pm$0.00 & 71.52$\pm$0.11 & 97.53$\pm$0.08 & \underline{97.57$\pm$0.00} & 96.25$\pm$0.02 & 97.44$\pm$0.00 & 96.85$\pm$0.02 & 96.78$\pm$0.12 & \cellcolor{oursgreen} \textbf{97.97$\pm$0.00} \\
\rowcolor{rowgray} DTD & 62.94$\pm$0.00 & 45.67$\pm$0.45 & 64.01$\pm$0.35 & 63.24$\pm$0.12 & 64.32$\pm$0.25 & 62.53$\pm$0.00 & 63.81$\pm$0.09 & \underline{64.34$\pm$0.24} & \cellcolor{oursgreen} \textbf{66.55$\pm$0.00} \\
EuroSAT & 43.79$\pm$0.00 & 23.31$\pm$0.04 & 27.05$\pm$0.14 & 42.92$\pm$11.45 & \underline{50.56$\pm$1.37} & 46.21$\pm$0.00 & 44.14$\pm$0.15 & 50.20$\pm$0.60 & \cellcolor{oursgreen} \textbf{52.41$\pm$0.00} \\
\rowcolor{rowgray} Aircraft & 43.91$\pm$0.02 & 18.52$\pm$0.28 & 35.46$\pm$0.39 & 41.36$\pm$0.18 & 45.51$\pm$0.57 & 41.40$\pm$0.00 & 46.23$\pm$0.13 & \underline{46.50$\pm$0.19} & \cellcolor{oursgreen} \textbf{47.07$\pm$0.00} \\
Flowers102 & 85.71$\pm$0.00 & 60.98$\pm$0.44 & 85.05$\pm$0.06 & 84.30$\pm$0.13 & \underline{86.51$\pm$0.55} & 85.91$\pm$0.00 & 84.76$\pm$0.05 & 86.26$\pm$0.22 & \cellcolor{oursgreen} \textbf{87.66$\pm$0.00} \\
\rowcolor{rowgray} Food101 & 86.82$\pm$0.00 & 40.45$\pm$0.10 & 88.85$\pm$0.03 & 88.61$\pm$0.08 & 87.04$\pm$0.05 & \textbf{89.50$\pm$0.00} & 86.85$\pm$0.07 & 86.88$\pm$0.05 & \cellcolor{oursgreen} \underline{89.22$\pm$0.00} \\
OxfordPets & 89.81$\pm$0.00 & 79.10$\pm$0.17 & 92.54$\pm$0.10 & 92.88$\pm$0.16 & 90.54$\pm$0.90 & \underline{93.73$\pm$0.00} & 89.03$\pm$0.06 & 90.16$\pm$0.12 & \cellcolor{oursgreen} \textbf{94.47$\pm$0.00} \\
\rowcolor{rowgray} StanfordCars & 90.70$\pm$0.00 & 74.90$\pm$0.23 & 91.17$\pm$0.10 & 91.28$\pm$0.14 & 90.83$\pm$0.09 & \underline{91.34$\pm$0.00} & 90.94$\pm$0.16 & 90.83$\pm$0.07 & \cellcolor{oursgreen} \textbf{91.57$\pm$0.00} \\
SUN397 & 67.41$\pm$0.00 & 50.38$\pm$0.13 & 68.22$\pm$0.04 & 68.81$\pm$0.28 & 68.92$\pm$0.09 & \underline{70.80$\pm$0.00} & 67.99$\pm$0.09 & 68.84$\pm$0.09 & \cellcolor{oursgreen} \textbf{72.09$\pm$0.00} \\
\rowcolor{rowgray} UCF101 & 65.42$\pm$0.00 & 38.35$\pm$0.12 & 70.20$\pm$0.24 & 71.83$\pm$0.25 & 66.82$\pm$0.45 & \underline{75.10$\pm$0.00} & 65.92$\pm$0.07 & 67.91$\pm$0.21 & \cellcolor{oursgreen} \textbf{79.01$\pm$0.00} \\
\midrule
\textbf{Mean} & 73.34 & 50.32 & 72.01 & 74.28 & 74.73 & \underline{75.40} & 73.65 & 74.87 & \cellcolor{oursgreen} \textbf{77.80} \\
\bottomrule
\end{tabular}
\end{adjustbox}

\vspace{2pt}
\begin{flushleft}
\footnotesize
\textit{Note.} Bold and underline indicate the best and second-best
methods, respectively.
\end{flushleft}
\vspace{-15pt}
\end{table*}

Tables~\ref{tab:vit_b16_fine_grained}, \ref{tab:vit_b32_fine_grained}, \ref{tab:vit_l14_fine_grained}, and \ref{tab:siglip_b16_fine_grained} report results on the fine-grained benchmark. Across all four backbones, \textsc{GATE} achieves the best mean accuracy. On SigLIP-B/16, \textsc{GATE} reaches 77.80\%, outperforming the strongest baseline, CARPRT, by 2.40 points. On CLIP ViT-B/16, it improves the mean accuracy from 67.35\% for the strongest baseline to 69.25\%. Similar trends hold for CLIP ViT-B/32 and CLIP ViT-L/14, where \textsc{GATE} improves over the strongest baselines by 2.31 and 1.39 points, respectively.

\textsc{GATE} is particularly effective on EuroSAT, OxfordPets, SUN397, and UCF101. On CLIP ViT-B/16, \textsc{GATE} improves over the strongest non-\textsc{GATE} method by 3.74 points on EuroSAT, 1.45 points on OxfordPets, 2.19 points on SUN397, and 1.77 points on UCF101. These gains suggest that image-side Gaussian evidence from reliable pseudo-evidence helps adapt the zero-shot classifier to target-domain structure.

Some datasets favor specialized baselines. CARPRT remains competitive on prompt-sensitive datasets such as DTD and Food101, while Zero and TPT are strong on some fine-grained settings. Thus, \textsc{GATE} achieves the strongest average performance rather than the best performance on every individual dataset.

\subsection{ImageNet-Family Distribution Shifts}


\begin{table*}[b]
\centering
\footnotesize
\setlength{\tabcolsep}{3.0pt}
\renewcommand{\arraystretch}{1.08}
\caption{
ImageNet-family distribution-shift results using CLIP ViT-L/14. Dataset rows report top-1 accuracy as mean$\pm$std over three random seeds; the Mean row reports the average across datasets.
}
\label{tab:vit_l14_imagenet_family}
\begin{adjustbox}{max width=\textwidth}
\begin{tabular}{l c| cccccccc}
\toprule
Dataset & Zero-shot & Zero & TPT & BITTA & TT-RAA & CARPRT & DN & TDA & \textsc{GATE} (ours) \\
\midrule
ImageNet & 72.84$\pm$0.00 & 74.89$\pm$0.07 & 75.53$\pm$0.09 & 73.80$\pm$0.57 & 73.82$\pm$0.08 & \underline{75.84$\pm$0.00} & 73.16$\pm$0.01 & 74.31$\pm$0.04 & \cellcolor{oursgreen} \textbf{77.70$\pm$0.06} \\
\rowcolor{rowgray} ImageNetA & 68.25$\pm$0.00 & \underline{76.99$\pm$0.19} & 76.21$\pm$0.18 & 72.92$\pm$0.20 & 69.28$\pm$0.09 & 71.56$\pm$0.00 & 68.84$\pm$0.09 & 69.67$\pm$0.15 & \cellcolor{oursgreen} \textbf{77.96$\pm$0.27} \\
ImageNetR & 85.20$\pm$0.00 & 88.13$\pm$0.10 & 87.69$\pm$0.06 & \underline{88.48$\pm$0.15} & 85.60$\pm$0.03 & 87.92$\pm$0.00 & 86.17$\pm$0.03 & 86.03$\pm$0.09 & \cellcolor{oursgreen} \textbf{89.95$\pm$0.07} \\
\rowcolor{rowgray} ImageNetSketch & 57.39$\pm$0.00 & \underline{59.90$\pm$0.10} & 59.41$\pm$0.03 & 57.74$\pm$0.10 & 58.74$\pm$0.06 & 59.28$\pm$0.00 & 58.19$\pm$0.04 & 59.53$\pm$0.07 & \cellcolor{oursgreen} \textbf{61.73$\pm$0.11} \\
\midrule
\textbf{Mean} & 70.92 & \underline{74.98} & 74.71 & 73.23 & 71.86 & 73.65 & 71.59 & 72.38 & \cellcolor{oursgreen} \textbf{76.84} \\
\bottomrule
\end{tabular}
\end{adjustbox}

\vspace{2pt}
\begin{flushleft}
\footnotesize
\textit{Note.} Bold and underline indicate the best and second-best
methods, respectively.
\end{flushleft}
\vspace{-15pt}
\end{table*}

Tables~\ref{tab:vit_b16_imagenet_family}, \ref{tab:vit_b32_imagenet_family}, and \ref{tab:vit_l14_imagenet_family} report results on the ImageNet-family benchmark. \textsc{GATE} achieves the best mean accuracy for all three CLIP backbones. On ViT-B/16, \textsc{GATE} improves the mean accuracy to 65.66\%, outperforming the strongest baseline by 2.01 points. On ViT-B/32, \textsc{GATE} reaches 57.08\%, improving over the strongest baseline by 1.74 points. On ViT-L/14, where the zero-shot classifier is already substantially stronger, \textsc{GATE} still improves the mean accuracy to 76.84\%, outperforming the strongest baseline by 1.86 points.

On ImageNet-A, \textsc{GATE} consistently improves over the strongest baseline across all three backbones. On ImageNet-R and ImageNet-Sketch, \textsc{GATE} also provides stable gains, indicating that the text Gaussian prior and image Gaussian target summary complement each other under style and rendition shifts. These results show that the proposed adaptation remains useful even when the underlying VLM backbone is strong.

\section{Ablation Studies}
\label{sec:ablations}

We analyze the main design choices of \textsc{GATE}. Unless otherwise specified, ablations use CLIP ViT-B/16 and are averaged over Caltech101, DTD, EuroSAT, FGVC-Aircraft, Food101, and Flowers102 under the same evaluation protocol as the main experiments. All main results use the fixed settings described in Sec.~\ref{sec:method}; the parameter sweeps below are diagnostic sensitivity analyses and are not used for dataset-specific tuning.

\definecolor{gatecolor}{RGB}{0,128,128}
\begin{figure}[t]
\centering

\begin{tikzpicture}[baseline=0pt,x=0.045cm,y=1.30cm]
    \draw[->] (0,0) -- (56,0);
    \draw[->] (0,0) -- (0,2.15);

    \foreach \x in {5,10,20,30,40,50} {
        \draw (\x,0) -- (\x,0) ++(0,-2pt);
        \node[font=\scriptsize,anchor=base,yshift=-9pt] at (\x,0) {\x};
    }

    \foreach \y/\lab in {0/64,1/65,2/66} {
        \draw (0,\y) -- (-1.4,\y);
        \node[left,font=\scriptsize] at (-1.4,\y) {\lab};
        \draw[gray!20] (0,\y) -- (54,\y);
    }

    \draw[gatecolor!80!black,thick]
        (5,0.46) -- (10,0.98) -- (20,1.30) --
        (30,0.80) -- (40,0.72) -- (50,0.70);

    \foreach \x/\y in {
        5/0.46,
        10/0.98,
        20/1.30,
        30/0.80,
        40/0.72,
        50/0.70
    } {
        \fill[gatecolor!80!black] (\x,\y) circle (1.6pt);
    }

    \node[font=\scriptsize,anchor=base,yshift=-22pt]
        at (27.5,0) {Evidence per class $m$};
    \node[rotate=90,font=\scriptsize]
        at (-14.5,1.05) {Accuracy (\%)};
\end{tikzpicture}%
\hspace{25pt}%
\begin{tikzpicture}[baseline=0pt,x=0.3cm,y=0.35cm]
    \draw[->] (0,0) -- (10.6,0);
    \draw[->] (0,0) -- (0,8.0);

    \foreach \x/\lab in {0/58,4/62,8/66} {
        \draw (\x,0) -- (\x,0) ++(0,-2pt);
        \node[font=\scriptsize,anchor=base,yshift=-9pt] at (\x,0) {\lab};
        \draw[gray!20] (\x,0) -- (\x,7.6);
    }

    \node[left,font=\scriptsize] at (0,7.2) {Zero-shot};
    \node[left,font=\scriptsize] at (0,5.4) {Text};
    \node[left,font=\scriptsize] at (0,3.6) {w/o rel.};
    \node[left,font=\scriptsize] at (0,1.8) {GATE};

    \filldraw[fill=gatecolor!45,draw=gatecolor!80!black]
        (0,6.8) rectangle (3.24,7.6);
    \filldraw[fill=gatecolor!45,draw=gatecolor!80!black]
        (0,5.0) rectangle (4.63,5.8);
    \filldraw[fill=gatecolor!45,draw=gatecolor!80!black]
        (0,3.2) rectangle (6.05,4.0);
    \filldraw[fill=gatecolor!45,draw=gatecolor!80!black]
        (0,1.4) rectangle (7.30,2.2);

    \node[right,font=\scriptsize] at (3.24,7.2) {61.24};
    \node[right,font=\scriptsize] at (4.63,5.4) {62.63};
    \node[right,font=\scriptsize] at (6.05,3.6) {64.05};
    \node[right,font=\scriptsize] at (7.30,1.8) {65.30};

    \node[font=\scriptsize,anchor=base,yshift=-22pt]
        at (5,0) {Accuracy (\%)};
\end{tikzpicture}

\vspace{4pt}
\makebox[0.48\linewidth]{\footnotesize\textbf{(a)} Evidence budget.}\hfill
\makebox[0.48\linewidth]{\footnotesize\textbf{(b)} Component analysis.}

\caption{
Ablation analysis of \textsc{GATE} on six datasets using CLIP ViT-B/16.
(a) Sensitivity to the pseudo-evidence budget $m$; the main method uses a fixed budget of up to 20 samples per class.
(b) Cumulative component analysis: zero-shot inference, text Gaussian evidence, image-side Gaussian evidence without reliability gating, and the full reliability-gated model.
}
\label{fig:ablation_summary}
\end{figure}
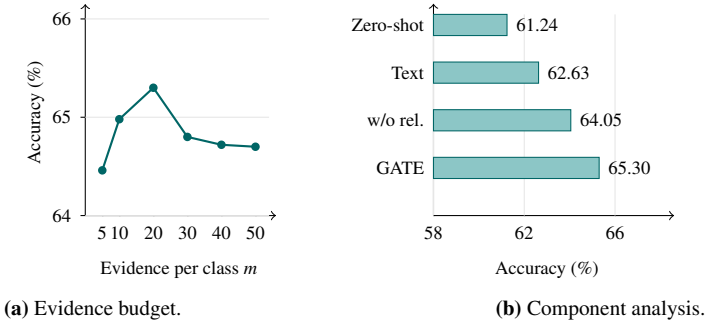

Figure~\ref{fig:ablation_summary}(a) studies the amount of pseudo-evidence retained for each class. With only a few target samples, the estimated image-side statistics are sensitive to individual pseudo-labels. Increasing the evidence budget improves accuracy up to the fixed budget of 20 samples per class, while retaining additional lower-confidence samples slightly reduces performance. This reflects a reliability--coverage trade-off, and we therefore use a fixed budget of up to 20 samples per class throughout the main experiments.

Figure~\ref{fig:ablation_summary}(b) evaluates the cumulative contribution of the main components. Zero-shot inference obtains $61.24\%$, while adding text Gaussian evidence increases accuracy to $62.63\%$. Adding image-side Gaussian evidence without reliability gating further improves performance to $64.05\%$, and the full reliability-gated \textsc{GATE} reaches $65.30\%$. Thus, the improvement is not attributable to a single component: language-derived semantic variation and target-domain image statistics provide complementary evidence, while reliability gating further limits the influence of weak or ambiguous pseudo-evidence.

\begin{figure}[!t]
\centering
\vspace{-0.6em}

\pgfplotsset{
  compat=1.16,
  fouraxis/.style={
    width=2.45cm,
    height=2.45cm,
    scale only axis,
    tick label style={font=\scriptsize},
    label style={font=\scriptsize},
    title style={font=\scriptsize},
    ylabel style={yshift=-6pt},
    grid=major,
    grid style={gray!18},
    major tick length=1pt,
    tick style={line width=0.25pt},
    legend image code/.code={
      \draw[#1] (0cm,0cm) -- (0.15cm,0cm);
    },
  }
}

\setlength{\tabcolsep}{0pt}

\begin{tabular}{
@{}c@{\hspace{0.05em}}
c@{\hspace{0.05em}}
c@{\hspace{0.05em}}
c@{}
}

\begin{tikzpicture}
\begin{axis}[
  fouraxis,
  title={(a) $\alpha,\rho$},
  xlabel={value},
  ylabel={Acc.\ (\%)},
  ymin=71.2,
  ymax=72.9,
  xmin=-0.03,
  xmax=0.95,
  xtick={0,0.4,0.8},
  ytick={71.5,72.0,72.5},
  legend columns=2,
  legend style={
    at={(0.5,1.02)},
    anchor=south,
    font=\scriptsize,
    draw=none,
    fill=none,
    inner sep=0.5pt,
    column sep=2pt
  },
  legend cell align=left,
]

\addlegendimage{
  thick,
  blue!70!black,
  mark=*,
  mark size=0.4pt
}
\addlegendentry{$\alpha$}

\addlegendimage{
  thick,
  brown!80!black,
  mark=square*,
  mark size=0.4pt
}
\addlegendentry{$\rho$}

\addplot[
  mark=*,
  mark size=0.4pt,
  thick,
  blue!70!black,
  forget plot
]
table[x=x,y=y]
{figure_sources/hp_alpha.dat};

\addplot[
  mark=square*,
  mark size=0.4pt,
  thick,
  brown!80!black,
  forget plot
]
table[x=x,y=y]
{figure_sources/hp_rho.dat};

\draw[
  blue!70!black,
  dashed,
  thick
]
(axis cs:0.10,71.2) -- (axis cs:0.10,72.9);

\draw[
  brown!80!black,
  dashed,
  thick
]
(axis cs:0.50,71.2) -- (axis cs:0.50,72.9);

\end{axis}
\end{tikzpicture}

&

\begin{tikzpicture}
\begin{axis}[
  fouraxis,
  title={(b) $\gamma{\times}\delta$},
  xlabel={$\delta$},
  ylabel={$\gamma$},
  enlargelimits=false,
  axis on top,
  xtick={0,1,2},
  xticklabels={.5,1,2},
  ytick={0,1,2},
  yticklabels={.5,1,2},
  colorbar horizontal,
  colormap/viridis,
  colorbar style={
    at={(0,1.04)},
    anchor=south west,
    width=2.45cm,
    height=0.05cm,
    xticklabel pos=upper,
    xtick={72.45,72.60},
    xticklabel style={font=\scriptsize},
    tick style={draw=none},
  },
  point meta min=72.44,
  point meta max=72.62,
]

\addplot[
  matrix plot*,
  mesh/cols=3,
  mesh/rows=3,
  point meta=explicit
]
table[x=di,y=gi,meta=acc]
{figure_sources/hp_gd.dat};

\end{axis}
\end{tikzpicture}

&

\begin{tikzpicture}
\begin{axis}[
  fouraxis,
  title={(c) corruption},
  xlabel={corr.\ (\%)},
  ylabel={Acc.\ (\%)},
  xmin=0,
  xmax=70,
  ymin=64,
  ymax=75,
  xtick={0,35,70},
  ytick={65,70,75},
  legend columns=2,
  legend style={
    at={(0.5,1.02)},
    anchor=south,
    font=\scriptsize,
    draw=none,
    fill=none,
    inner sep=0.5pt,
    column sep=2pt
  },
  legend cell align=left,
]

\addlegendimage{
  thick,
  blue!60!black,
  mark=*,
  mark size=0.4pt
}
\addlegendentry{\method}

\addlegendimage{red,dashed,thick}
\addlegendentry{zero-shot}

\addplot[
  mark=*,
  mark size=0.4pt,
  thick,
  blue!60!black,
  forget plot
]
table[x=x,y=y]
{figure_sources/corrupt.dat};

\addplot[
  red,
  dashed,
  thick,
  domain=0:70,
  forget plot
]
{65.64};

\end{axis}
\end{tikzpicture}

&

\begin{tikzpicture}
\begin{axis}[
  fouraxis,
  title={(d) starvation},
  xlabel={starv.\ (\%)},
  xmin=0,
  xmax=70,
  ymin=64,
  ymax=75,
  xtick={0,35,70},
  ytick={65,70,75},
  yticklabels={},
  legend columns=2,
  legend style={
    at={(0.5,1.02)},
    anchor=south,
    font=\scriptsize,
    draw=none,
    fill=none,
    inner sep=0.5pt,
    column sep=2pt
  },
  legend cell align=left,
]

\addlegendimage{
  thick,
  blue!60!black,
  mark=*,
  mark size=0.4pt
}
\addlegendentry{\method}

\addlegendimage{red,dashed,thick}
\addlegendentry{zero-shot}

\addplot[
  mark=*,
  mark size=0.4pt,
  thick,
  blue!60!black,
  forget plot
]
table[x=x,y=y]
{figure_sources/starve.dat};

\addplot[
  red,
  dashed,
  thick,
  domain=0:70,
  forget plot
]
{65.64};

\end{axis}
\end{tikzpicture}

\end{tabular}
\caption{\small
Sensitivity and pseudo-evidence robustness on UCF101 with CLIP ViT-B/16.
(a) Residual-strength and image-covariance-shrinkage sweeps; dashed lines indicate the fixed settings.
(b) Sensitivity to the evidence-count and reliability exponents.
(c) Robustness to corrupted pseudo-evidence.
(d) Robustness to evidence starvation.
}
\label{fig:sensitivity_robustness}

\end{figure}
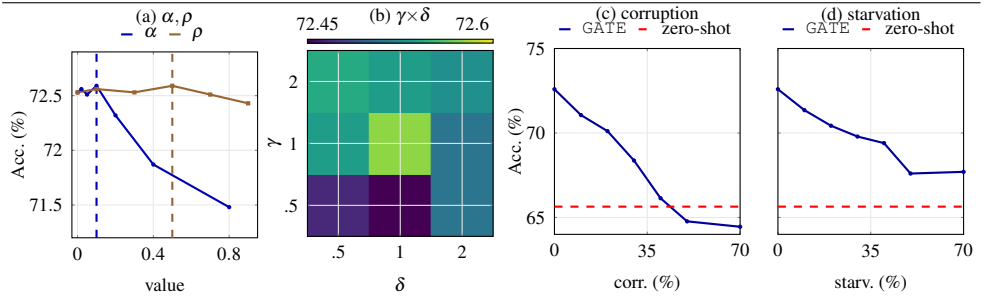

Figure~\ref{fig:sensitivity_robustness}(a,b) evaluates sensitivity to the main correction and gating parameters on UCF101 with CLIP ViT-B/16. Here, $\alpha$ is the residual strength in Eq.~\eqref{eq:residual_correction}, $\rho$ is the image-covariance shrinkage coefficient in Eq.~\eqref{eq:image_covariance}, and $\gamma$ and $\delta$ are the evidence-count and confidence exponents in Eq.~\eqref{eq:reliability_gate}, respectively. The main method uses $\alpha=0.10$, $\rho=0.5$, and $\gamma=\delta=1$. Accuracy changes by only $0.16$ percentage points over $\rho\in\{0,0.1,0.3,0.5,0.7,0.9\}$ and by only $0.15$ points when $\gamma,\delta\in\{0.5,1,2\}$. Together with the residual-strength sweep, these results exhibit broad performance plateaus rather than narrow optima. All sweeps are diagnostic only; the fixed main settings are reused unchanged throughout the full benchmark.

Figure~\ref{fig:sensitivity_robustness}(c,d) examines robustness to degraded pseudo-evidence. Redirecting Pass-1 pseudo-evidence to randomly selected incorrect classes reduces \textsc{GATE} from $72.59\%$ to $66.14\%$ at $40\%$ corruption, while the frozen zero-shot classifier obtains $65.64\%$. We separately simulate missing or imbalanced evidence by restricting an increasing fraction of classes to one Pass-1 evidence sample. Even when $70\%$ of classes are evidence-starved, \textsc{GATE} obtains $67.70\%$, compared with $65.64\%$ for zero-shot inference. These results show that the reliability gate and bounded residual correction mitigate, but do not eliminate, dependence on pseudo-evidence quality and coverage.


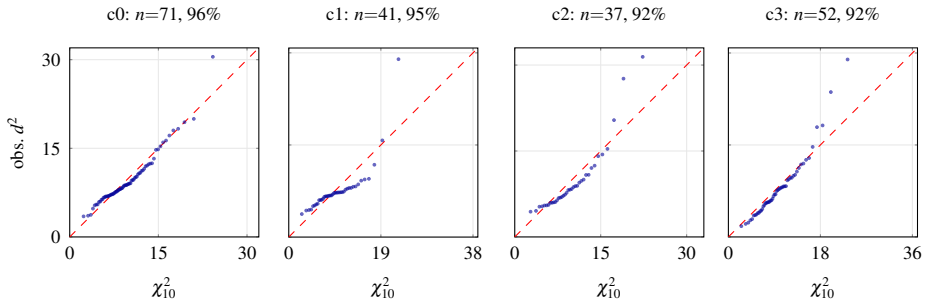
\begin{figure}[!t]
\centering

\pgfplotsset{
  width=2.5cm,
  height=2.5cm,
  scale only axis,
  tick label style={font=\scriptsize},
  label style={font=\scriptsize},
  title style={font=\scriptsize},
  grid=major,
  grid style={gray!18},
  major tick length=1pt,
  tick style={line width=0.25pt},
}

\setlength{\tabcolsep}{1pt}

\begin{tabular}{cccc}

\begin{tikzpicture}
\begin{axis}[
  xmin=0,xmax=32,
  ymin=0,ymax=32,
  xlabel={$\chi^2_{10}$},
  ylabel={obs.\ $d^2$},
  title={c0: $n{=}71$, 96\%},
  xtick={0,15,30},
  ytick={0,15,30}
]
\addplot[
  only marks,
  mark size=0.5pt,
  opacity=0.6,
  blue!60!black
]
table {figure_sources/qq_class0.dat};

\addplot[
  red,
  dashed,
  domain=0:32,
  samples=2
]
{x};
\end{axis}
\end{tikzpicture}

&

\begin{tikzpicture}
\begin{axis}[
  xmin=0,xmax=39,
  ymin=0,ymax=39,
  xlabel={$\chi^2_{10}$},
  title={c1: $n{=}41$, 95\%},
  xtick={0,19,38},
  ytick={0,19,38},
  yticklabels={}
]
\addplot[
  only marks,
  mark size=0.5pt,
  opacity=0.6,
  blue!60!black
]
table {figure_sources/qq_class1.dat};

\addplot[
  red,
  dashed,
  domain=0:39,
  samples=2
]
{x};
\end{axis}
\end{tikzpicture}

&

\begin{tikzpicture}
\begin{axis}[
  xmin=0,xmax=33,
  ymin=0,ymax=33,
  xlabel={$\chi^2_{10}$},
  title={c2: $n{=}37$, 92\%},
  xtick={0,15,30},
  ytick={0,15,30},
  yticklabels={}
]
\addplot[
  only marks,
  mark size=0.5pt,
  opacity=0.6,
  blue!60!black
]
table {figure_sources/qq_class2.dat};

\addplot[
  red,
  dashed,
  domain=0:33,
  samples=2
]
{x};
\end{axis}
\end{tikzpicture}

&

\begin{tikzpicture}
\begin{axis}[
  xmin=0,xmax=37,
  ymin=0,ymax=37,
  xlabel={$\chi^2_{10}$},
  title={c3: $n{=}52$, 92\%},
  xtick={0,18,36},
  ytick={0,18,36},
  yticklabels={}
]
\addplot[
  only marks,
  mark size=0.5pt,
  opacity=0.6,
  blue!60!black
]
table {figure_sources/qq_class3.dat};

\addplot[
  red,
  dashed,
  domain=0:37,
  samples=2
]
{x};
\end{axis}
\end{tikzpicture}

\end{tabular}

\caption{
Gaussian compatibility diagnostic on UCF101 with CLIP ViT-B/16.
Per-class Mahalanobis QQ plots are computed in a top-10 PCA diagnostic subspace against the $\chi^2_{10}$ reference.
Across the four representative classes, $92$--$96\%$ of samples lie within the theoretical $95\%$ envelope, with Henze--Zirkler statistics of $0.02$--$0.04$.
The remaining deviations are concentrated mainly in the upper tail and are mitigated by reliability gating.
}
\label{fig:gauss_qq}

\end{figure}

These diagnostics support using Gaussian statistics as regularized
second-order compatibility summaries rather than as full generative
models of VLM features.


\noindent\textbf{Limitations.}
\textsc{GATE} is transductive: Pass~1 estimates statistics from the full
unlabeled target split and Pass~2 predicts. Its performance still depends
on pseudo-evidence quality and coverage, although reliability gating and
the bounded residual reduce this sensitivity. It may underperform
prompt-specialized methods when the dominant error is semantic rather
than visual; accordingly, we claim the strongest average, not universal
per-dataset, performance.

\section{Conclusion}
We presented \texttt{GATE}, a training-free two-pass transductive method
that fuses text and target-image Gaussian evidence through class-wise
reliability gating without updating the VLM. The text Gaussian provides
a semantic prior from multiple language descriptions, while the image
Gaussian summarizes target-domain structure from reliable unlabeled
samples. Language-derived semantic variation and image-derived target
statistics provide complementary evidence. Across fine-grained and
ImageNet-family shifts with CLIP/SigLIP backbones, \texttt{GATE}
consistently improves zero-shot classification and achieves the strongest
average performance.



\section*{Acknowledgments}
This work was supported in part by a gift from The BMW Group.





\bibliography{ref}

@article{farina2024frustratingly,
  title={Frustratingly easy test-time adaptation of vision-language models},
  author={Farina, Matteo and Franchi, Gianni and Iacca, Giovanni and Mancini, Massimiliano and Ricci, Elisa},
  journal={Advances in Neural Information Processing Systems},
  volume={37},
  pages={129062--129093},
  year={2024}
}

@inproceedings{karmanov2024efficient,
  title={Efficient test-time adaptation of vision-language models},
  author={Karmanov, Adilbek and Guan, Dayan and Lu, Shijian and El Saddik, Abdulmotaleb and Xing, Eric},
  booktitle={Proceedings of the IEEE/CVF Conference on Computer Vision and Pattern Recognition},
  pages={14162--14171},
  year={2024}
}

@article{zhou2023test,
  title={Test-time distribution normalization for contrastively learned visual-language models},
  author={Zhou, Yifei and Ren, Juntao and Li, Fengyu and Zabih, Ramin and Lim, Ser Nam},
  journal={Advances in Neural Information Processing Systems},
  volume={36},
  pages={47105--47123},
  year={2023}
}

@inproceedings{sun2026bilateral,
  title={Bilateral information-aware test-time adaptation for vision-language models},
  author={Sun, Jingwei and Zhu, Jianing and Yao, Jiangchao and Niu, Gang and Sugiyama, Masashi and Han, Bo},
  booktitle={The Fourteenth International Conference on Learning Representations},
  year={2026}
}

@inproceedings{fan2025test,
  title={Test-Time Retrieval-Augmented Adaptation for Vision-Language Models},
  author={Fan, Xinqi and Chen, Xueli and Yang, Luoxiao and Yap, Chuin Hong and Qureshi, Rizwan and Dou, Qi and Yap, Moi Hoon and Shah, Mubarak},
  booktitle={Proceedings of the IEEE/CVF International Conference on Computer Vision},
  pages={8810--8819},
  year={2025}
}

@inproceedings{dong2026carprt,
  title = {{CARPRT}: Class-Aware Zero-Shot Prompt Reweighting for Black-Box Vision-Language Models},
  author    = {Dong, Ruijiang and Ye, Zesheng and Qi, Jianzhong and Feng, Lei and Liu, Feng and Niu, Gang and Sugiyama, Masashi},
  booktitle = {International Conference on Learning Representations (ICLR)},
  year      = {2026}
}

@inproceedings{radford2021learning,
  title={Learning transferable visual models from natural language supervision},
  author={Radford, Alec and Kim, Jong Wook and Hallacy, Chris and Ramesh, Aditya and Goh, Gabriel and Agarwal, Sandhini and Sastry, Girish and Askell, Amanda and Mishkin, Pamela and Clark, Jack and others},
  booktitle={International conference on machine learning},
  pages={8748--8763},
  year={2021},
  organization = {{PMLR}}
}

@inproceedings{lin2022frozen,
  title = {Frozen {CLIP} Models Are Efficient Video Learners},
  author={Lin, Ziyi and Geng, Shijie and Zhang, Renrui and Gao, Peng and De Melo, Gerard and Wang, Xiaogang and Dai, Jifeng and Qiao, Yu and Li, Hongsheng},
  booktitle={European Conference on Computer Vision},
  pages={388--404},
  year={2022},
  organization={Springer}
}

@inproceedings{guzhov2022audioclip,
  title = {{AudioCLIP}: Extending {CLIP} to Image, Text and Audio},
  author={Guzhov, Andrey and Raue, Federico and Hees, J{\"o}rn and Dengel, Andreas},
  booktitle={ICASSP 2022-2022 IEEE International Conference on Acoustics, Speech and Signal Processing (ICASSP)},
  pages={976--980},
  year={2022},
  organization={IEEE}
}

@inproceedings{liu2023clip,
  title = {{CLIP}-Driven Universal Model for Organ Segmentation and Tumor Detection},
  author={Liu, Jie and Zhang, Yixiao and Chen, Jie-Neng and Xiao, Junfei and Lu, Yongyi and A Landman, Bennett and Yuan, Yixuan and Yuille, Alan and Tang, Yucheng and Zhou, Zongwei},
  booktitle={Proceedings of the IEEE/CVF international conference on computer vision},
  pages={21152--21164},
  year={2023}
}

@article{li2021align,
  title={Align before fuse: Vision and language representation learning with momentum distillation},
  author={Li, Junnan and Selvaraju, Ramprasaath and Gotmare, Akhilesh and Joty, Shafiq and Xiong, Caiming and Hoi, Steven Chu Hong},
  journal={Advances in neural information processing systems},
  volume={34},
  pages={9694--9705},
  year={2021}
}

@inproceedings{li2023blip,
  title = {{BLIP}-2: Bootstrapping Language-Image Pre-Training with Frozen Image Encoders and Large Language Models},
  author={Li, Junnan and Li, Dongxu and Savarese, Silvio and Hoi, Steven},
  booktitle={International conference on machine learning},
  pages={19730--19742},
  year={2023},
  organization={PMLR}
}

@article{zeng2023x,
  author  = {Zeng, Yan and Zhang, Xinsong and Li, Hang and Wang, Jiawei
             and Zhang, Jipeng and Zhou, Wangchunshu},
  title   = {{$X^2$-VLM}: All-in-One Pre-Trained Model for
             Vision-Language Tasks},
  journal = {IEEE Transactions on Pattern Analysis and Machine
             Intelligence},
  volume  = {46},
  number  = {5},
  pages   = {3156--3168},
  year    = {2024},
  doi     = {10.1109/TPAMI.2023.3339661}
}

@article{ma2023swapprompt,
  title={Swapprompt: Test-time prompt adaptation for vision-language models},
  author={Ma, Xiaosong and Zhang, Jie and Guo, Song and Xu, Wenchao},
  journal={Advances in Neural Information Processing Systems},
  volume={36},
  pages={65252--65264},
  year={2023}
}

@inproceedings{lee2025ra,
  title = {{RA-TTA}: Retrieval-Augmented Test-Time Adaptation for Vision-Language Models},
  author={Lee, Youngjun and Kim, Doyoung and Kang, Junhyeok and Bang, Jihwan and Song, Hwanjun and Lee, Jae-Gil},
  booktitle={The Thirteenth International Conference on Learning Representations},
  year={2025}
}

@inproceedings{feng2023diverse,
  title={Diverse data augmentation with diffusions for effective test-time prompt tuning},
  author={Feng, Chun-Mei and Yu, Kai and Liu, Yong and Khan, Salman and Zuo, Wangmeng},
  booktitle={Proceedings of the IEEE/CVF International Conference on Computer Vision},
  pages={2704--2714},
  year={2023}
}

@inproceedings{zhai2023sigmoid,
  title={Sigmoid loss for language image pre-training},
  author={Zhai, Xiaohua and Mustafa, Basil and Kolesnikov, Alexander and Beyer, Lucas},
  booktitle={Proceedings of the IEEE/CVF international conference on computer vision},
  pages={11975--11986},
  year={2023}
}

@article{shu2022tpt,
  title={Test-time prompt tuning for zero-shot generalization in vision-language models},
  author={Shu, Manli and Nie, Weili and Huang, De-An and Yu, Zhiding and Goldstein, Tom and Anandkumar, Anima and Xiao, Chaowei},
  journal={Advances in Neural Information Processing Systems},
  volume={35},
  pages={14274--14289},
  year={2022}
}

@inproceedings{fei2004learning,
  author    = {Li, Fei-Fei and Fergus, Rob and Perona, Pietro},
  title     = {Learning Generative Visual Models from Few Training
               Examples: An Incremental Bayesian Approach Tested on
               101 Object Categories},
  booktitle = {2004 Conference on Computer Vision and Pattern
               Recognition Workshop},
  pages     = {178--178},
  year      = {2004},
  publisher = {IEEE}
}

@inproceedings{cimpoi2014describing,
  author    = {Cimpoi, Mircea and Maji, Subhransu and Kokkinos, Iasonas
               and Mohamed, Sammy and Vedaldi, Andrea},
  title     = {Describing Textures in the Wild},
  booktitle = {Proceedings of the IEEE Conference on Computer Vision
               and Pattern Recognition},
  pages     = {3606--3613},
  year      = {2014}
}

@article{helber2019eurosat,
  author  = {Helber, Patrick and Bischke, Benjamin and Dengel, Andreas
             and Borth, Damian},
  title   = {{EuroSAT}: A Novel Dataset and Deep Learning Benchmark for
             Land Use and Land Cover Classification},
  journal = {IEEE Journal of Selected Topics in Applied Earth
             Observations and Remote Sensing},
  volume  = {12},
  number  = {7},
  pages   = {2217--2226},
  year    = {2019},
  doi     = {10.1109/JSTARS.2019.2918242}
}

@techreport{maji2013fine,
  author        = {Maji, Subhransu and Rahtu, Esa and Kannala, Juho
                   and Blaschko, Matthew and Vedaldi, Andrea},
  title         = {Fine-Grained Visual Classification of Aircraft},
  institution   = {arXiv},
  year          = {2013},
  archivePrefix = {arXiv},
  eprint        = {1306.5151},
  primaryClass  = {cs.CV}
}

@inproceedings{nilsback2008automated,
  author    = {Nilsback, Maria-Elena and Zisserman, Andrew},
  title     = {Automated Flower Classification over a Large Number of
               Classes},
  booktitle = {2008 Sixth Indian Conference on Computer Vision,
               Graphics and Image Processing},
  pages     = {722--729},
  year      = {2008},
  publisher = {IEEE},
  doi       = {10.1109/ICVGIP.2008.47}
}

@inproceedings{bossard2014food,
  author    = {Bossard, Lukas and Guillaumin, Matthieu and
               Van Gool, Luc},
  title     = {{Food-101}: Mining Discriminative Components with
               Random Forests},
  booktitle = {Computer Vision -- ECCV 2014},
  pages     = {446--461},
  year      = {2014},
  publisher = {Springer}
}

@inproceedings{parkhi2012cats,
  author    = {Parkhi, Omkar M. and Vedaldi, Andrea and
               Zisserman, Andrew and Jawahar, C. V.},
  title     = {Cats and Dogs},
  booktitle = {2012 IEEE Conference on Computer Vision and
               Pattern Recognition},
  pages     = {3498--3505},
  year      = {2012},
  publisher = {IEEE},
  doi       = {10.1109/CVPR.2012.6248092}
}

@inproceedings{krause2013object,
  author    = {Krause, Jonathan and Stark, Michael and Deng, Jia
               and Li, Fei-Fei},
  title     = {{3D} Object Representations for Fine-Grained
               Categorization},
  booktitle = {2013 IEEE International Conference on Computer Vision
               Workshops},
  pages     = {554--561},
  year      = {2013},
  publisher = {IEEE},
  doi       = {10.1109/ICCVW.2013.77}
}

@inproceedings{xiao2010sun,
  author    = {Xiao, Jianxiong and Hays, James and Ehinger, Krista A.
               and Oliva, Aude and Torralba, Antonio},
  title     = {{SUN} Database: Large-Scale Scene Recognition from
               Abbey to Zoo},
  booktitle = {2010 IEEE Computer Society Conference on Computer
               Vision and Pattern Recognition},
  pages     = {3485--3492},
  year      = {2010},
  publisher = {IEEE},
  doi       = {10.1109/CVPR.2010.5539970}
}

@article{soomro2012ucf101,
  author        = {Soomro, Khurram and Zamir, Amir Roshan and
                   Shah, Mubarak},
  title         = {{UCF101}: A Dataset of 101 Human Actions Classes
                   From Videos in the Wild},
  journal       = {arXiv preprint arXiv:1212.0402},
  year          = {2012}
}

@inproceedings{deng2009imagenet,
  author    = {Deng, Jia and Dong, Wei and Socher, Richard and
               Li, Li-Jia and Li, Kai and Li, Fei-Fei},
  title     = {{ImageNet}: A Large-Scale Hierarchical Image Database},
  booktitle = {2009 IEEE Conference on Computer Vision and
               Pattern Recognition},
  pages     = {248--255},
  year      = {2009},
  publisher = {IEEE},
  doi       = {10.1109/CVPR.2009.5206848}
}

@inproceedings{hendrycks2021natural,
  author    = {Hendrycks, Dan and Zhao, Kevin and Basart, Steven and
               Steinhardt, Jacob and Song, Dawn},
  title     = {Natural Adversarial Examples},
  booktitle = {Proceedings of the IEEE/CVF Conference on Computer
               Vision and Pattern Recognition},
  pages     = {15262--15271},
  year      = {2021}
}

@inproceedings{hendrycks2021many,
  author    = {Hendrycks, Dan and Basart, Steven and Mu, Norman and
               Kadavath, Saurav and Wang, Frank and Dorundo, Evan and
               Desai, Rahul and Zhu, Tyler and Parajuli, Samyak and
               Guo, Mike and Song, Dawn and Steinhardt, Jacob and
               Gilmer, Justin},
  title     = {The Many Faces of Robustness: A Critical Analysis of
               Out-of-Distribution Generalization},
  booktitle = {Proceedings of the IEEE/CVF International Conference
               on Computer Vision},
  pages     = {8340--8349},
  year      = {2021}
}

@inproceedings{wang2019learning,
  author    = {Wang, Haohan and Ge, Songwei and Lipton, Zachary C.
               and Xing, Eric P.},
  title     = {Learning Robust Global Representations by Penalizing
               Local Predictive Power},
  booktitle = {Advances in Neural Information Processing Systems},
  volume    = {32},
  pages     = {10506--10518},
  year      = {2019}
}

\end{document}